\documentclass[letterpaper, 10 pt, conference]{ieeeconf}  

\IEEEoverridecommandlockouts                              

\usepackage{graphics} 
\usepackage{epsfig} 
\usepackage{mathptmx} 
\usepackage{mathrsfs}
\usepackage{times} 
\usepackage{amsmath} 
\usepackage{amssymb}  
\usepackage{cite}
\usepackage{times}
\usepackage{mathtools}
\usepackage{multicol}
\usepackage[bookmarks=true]{hyperref}
\usepackage{cleveref}
\usepackage{graphicx}
\usepackage{array,multirow,booktabs} 
\graphicspath{{figures/}}
\usepackage{caption}
\usepackage{subcaption}
\usepackage{amsmath,amssymb} 
\usepackage{pifont}  
\usepackage{afterpage}
\usepackage[dvipsnames]{xcolor}
\usepackage{xspace}
\newcommand{\eg}{\textit{e.g.}\xspace}
\newcommand{\ie}{\textit{i.e.}\xspace}

\title{\Large \bf \textit{DemoDiffusion}: One-Shot Human Imitation using pre-trained Diffusion Policy}

\author{Sungjae Park, Homanga Bharadhwaj, Shubham Tulsiani
\\ Carnegie Mellon University
}

\newcommand{\approach}{\text{\textit{DemoDiffusion}}}

\begin{document}
\makeatletter

\def\@maketitle{%
  \newpage
  \null
  \vskip 0.5em 
  \begin{center}%
    {\LARGE \bf \@title \par}%
    \vskip 0.7em 
    {\large
      \lineskip .5em%
      \begin{tabular}[t]{c}%
        \@author
      \end{tabular}\par}%
  \end{center}%
  \par
  \vskip 0.3em 
}

\let\@oldmaketitle\@maketitle%
\renewcommand{\@maketitle}{\@oldmaketitle%
\includegraphics[width=\linewidth]{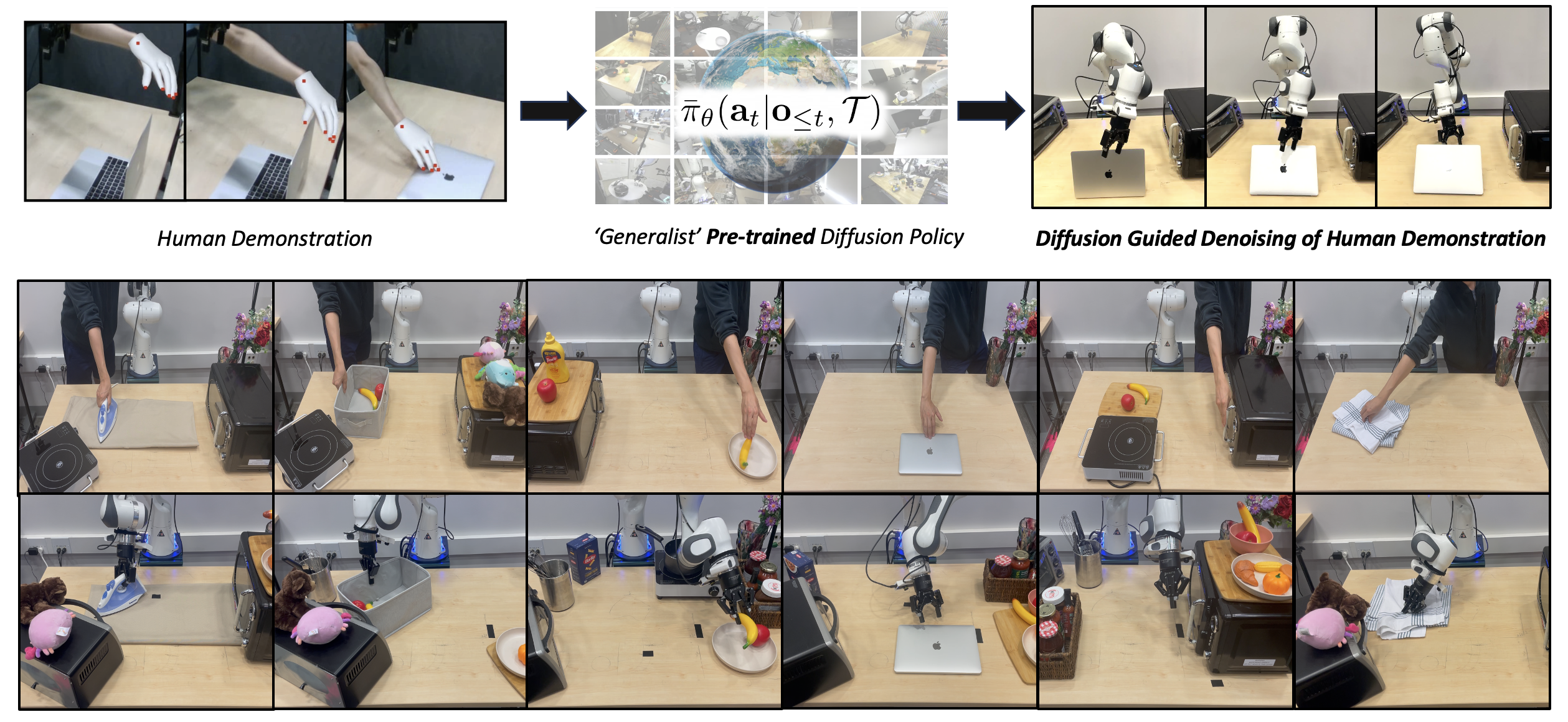} 
\refstepcounter{figure}{{Fig. \thefigure:} \textbf{Overview of~\approach.} We show how \textit{generalist} pre-trained diffusion policies can be used for following a generic human demonstration showing a manipulation task during deployment. Our real-world manipulation results encompass a wide diversity of manipulation tasks involving everyday objects.}
\label{fig:teaser}
\medskip
\vspace{-0.3cm}}%

\makeatother

\maketitle
\thispagestyle{empty}
\pagestyle{empty}

\renewcommand\thefigure{\arabic{figure}}
\setcounter{figure}{1}

\begin{abstract}
We propose \approach, a simple method for enabling robots to perform manipulation tasks by imitating a single human demonstration, without requiring task-specific training or paired human-robot data. Our approach is based on two insights. First, the hand motion in a human demonstration provides a useful prior for the robot’s end-effector trajectory, which we can convert into a rough open-loop robot motion trajectory via kinematic retargeting. Second, while this retargeted motion captures the overall structure of the task, it may not align well with plausible robot actions in-context. To address this, we leverage a pre-trained \textit{generalist} diffusion policy to modify the trajectory, ensuring it both follows the human motion and remains within the distribution of plausible robot actions. Unlike approaches based on online reinforcement learning or paired human-robot data, our method enables robust adaptation to new tasks and scenes with minimal effort. In real-world experiments across 8 diverse manipulation tasks, DemoDiffusion achieves 83.8\% average success rate, compared to 13.8\% for the pre-trained policy and 52.5\% for kinematic retargeting, succeeding even on tasks where the pre-trained \textit{generalist} policy fails entirely. Project page:
\noindent\textcolor{purple}{\url{https://demodiffusion.github.io/}}
\end{abstract}


\section{Introduction}
\label{sec:introduction}

How do we build robot manipulation systems that can be readily deployed in unstructured human environments? One possible answer is to learn `generalist' policies that are capable of accomplishing any generic task (specified via some description \eg language or image goal) in any environment. Indeed, there is a general optimism about this paradigm, reflected in several ongoing efforts to collect large-scale demonstration datasets to train such policies. While these efforts have produced unified policies capable of diverse tasks~\cite{pi0,rt2,bcz, openvla, rt1}, these policies still struggle to perform meaningfully when deployed zero-shot to novel environments or asked to perform unseen tasks. Deployment thus often requires additional fine-tuning using task-and-scene-specific robot demonstrations~\cite{pi0}, but this is a non-trivial overhead as collecting robot demonstrations in the real world can be time-consuming and beyond the expertise of an average user. In this work, we propose an alternate deployment mechanism -- leveraging a pre-trained generalist policy to perform a task via imitating a single \emph{human} demonstration.

We are of course not the first to consider this goal of allowing robots to imitate any given human demonstration. One common approach~\cite{li2024okami, r+x, wang2024dexcap, chen2024arcap} is to instantiate this as a kinematic retargeting task and compute open-loop robot actions that maximize a manually defined similarity between the achieved robot end-effector configuration and observed human hand poses (\eg matching locations of fingertips). However, the human-robot embodiment mismatch makes it difficult for the retargeted actions to achieve precisely the same effects as the human ones, and the open-loop execution further makes the approach brittle to noise and scene variations. Another line of work for human imitation attempts to learn robot policies through online reinforcement learning~\cite{bahl2022human,irmak, chen2024object, liu2025dextrack, liu2024quasisim}, where the human demonstration helps define reward functions. While this can overcome the embodiment gap, such test-time online RL requires hours of online interaction and resets, making it difficult to adopt for generic real-world manipulation tasks, especially in safety-critical scenarios. We seek an approach that, like kinematic retargeting, can be deployed one-shot (\ie without test-time training), but still benefits from learned priors for precise closed-loop interaction.

Our approach builds on the insight that pre-trained diffusion policies can act as priors for robot action. Inspired by prior work in leveraging pre-trained diffusion models for image editing~\cite{meng2021sdedit}, we 
present~\approach~-- a formulation to utilize a diffusion policy trained on robot interaction data for synthesizing coherent robot actions from a human demonstration. Specifically, we first perform kinematic retargeting by extracting human hand poses from the demonstration and obtain an open-loop robot action trajectory. While this trajectory is typically suboptimal due to embodiment differences and lack of closed-loop feedback, it serves as an effective initialization that can be improved via a diffusion policy. We do so by injecting Gaussian noise and applying the pre-trained diffusion policy to iteratively denoise the trajectory conditioned on robot observations, yielding a refined, executable sequence of robot actions. \approach~thus enables the robot to use the pre-trained policy as a prior and adapt the human-derived trajectory to its own embodiment and environment in a closed-loop manner.

In summary, we propose~\approach: a framework for robotic manipulation that allows a robot to perform generic tasks in natural environments by following a human demonstration with guidance from a pre-trained diffusion policy. 
Importantly, our framework doesn't require any robot demonstration of the target task within the target environment, online interaction, or fine-tuning. Our experiments across simulation and real-world environments show that~\approach~surpasses both the base policy and retargeted trajectory, achieving 83.8\% success across 8 real-world tasks -- even those where the pre-trained \textit{generalist} policy fails entirely.

\section{Related Work}
\label{sec:related_work}
\begin{figure*}[t!]
\centering
\includegraphics[width=\textwidth]{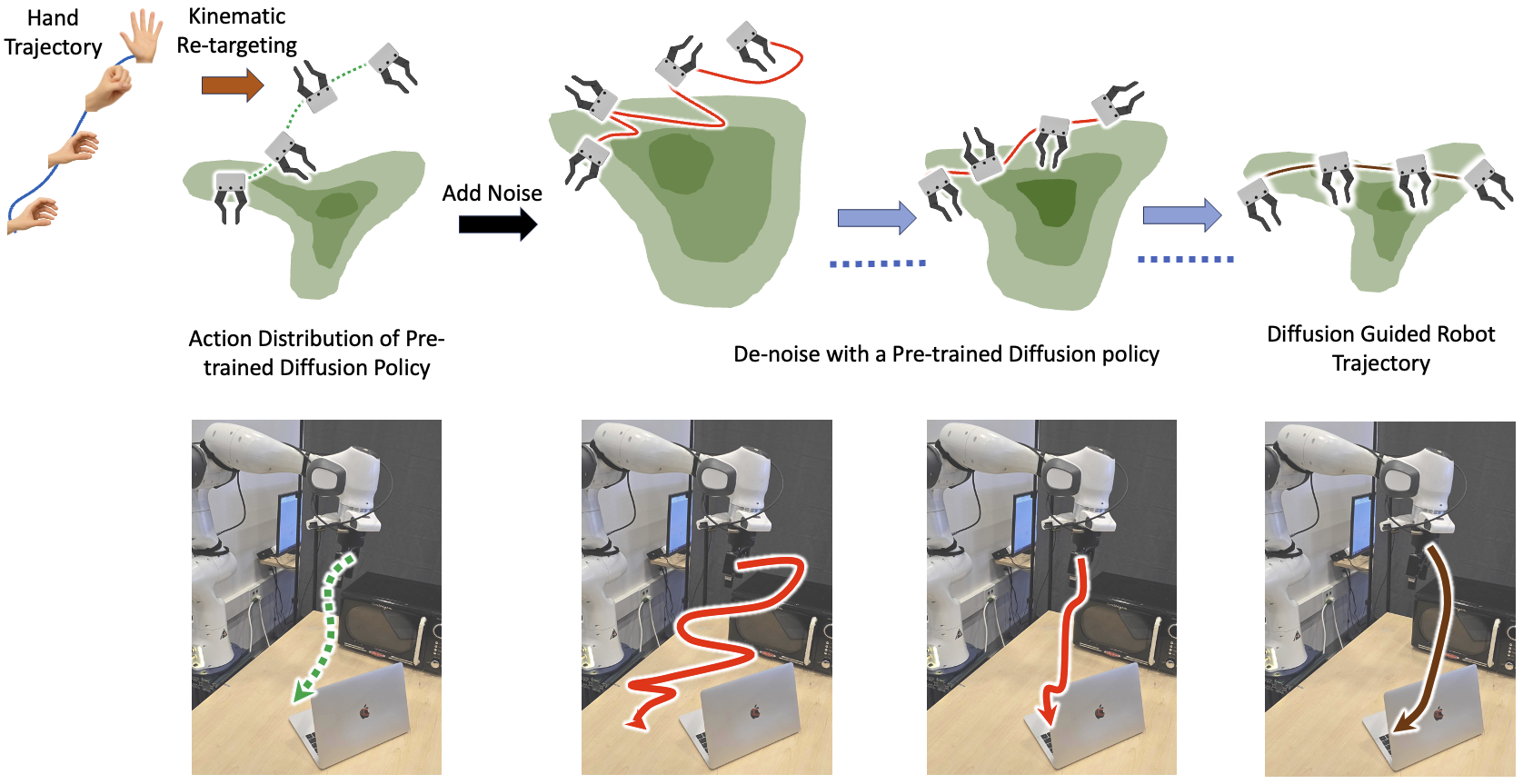}
    \caption{\textbf{Retargeted human hand trajectory to closed-loop robot action sequence, for the task $\mathcal{T}$: \textit{``shut down the laptop"}}. The dotted line shows the trajectory of robot end-effector poses after kinematic retargeting. The olive contour plot depicts the distribution of trajectories from a pre-trained diffusion policy. Given a kinematic retargeting, we first perturb it with Gaussian noise and progressively remove the noise by simulating the reverse SDE with the diffusion policy. This process gradually projects a potentially unfeasible but approximately correct retargeting to the manifold of plausible robot actions that can perform real-world manipulation, in this case closing the laptop without missing the edge.}

\label{fig:intuition}
\end{figure*} 
\textbf{{\textit{Generalist} Manipulation Policies.}} There has been a growing trend in developing `\textit{generalist}' robotic policies that can perform multiple tasks based on a specified goal in the form of an image or a language instruction. Collecting large-scale robot interaction datasets~\cite{rh20t,walke2023bridgedata} combined with behavior cloning and vision-language pre-training~\cite{rt1,rt2,roboagent,pi0,openvla,bcz, intelligence2025pi_} is the predominant recipe for training such multi-task policies. However, since collecting large-scale robot interaction datasets in natural settings like homes and offices is challenging due to operational constraints, current best models still struggle to perform manipulation tasks zero-shot and thus are not yet deployable in-the-wild. Our work enables using such \textit{generalist} diffusion policies for following a human demonstration, enabling it to perform tasks that the pre-trained policy struggles to zero-shot.

\textbf{Robotic Manipulation with Non-Robot Datasets.}  Instead of requiring large-scale robot interaction datasets, a growing body of works have begun utilizing  human videos and large-scale web videos for robotics. These approaches have been enabled by recent advances in computer vision for representation learning~\cite{he2022masked, dosovitskiy2020image, zhu2025large}, predicting tracks~\cite{cotracker,tapir}, and reconstructing hand-object interactions from monocular videos~\cite{rong2021frankmocap,pavlakos2024reconstructing,ye2022s, ye2023diffusion, ye2023diffusion}. Earlier works learned self-supervised visual representations from such non-robotic datasets~\cite{damen2018scaling,ego4d, grauman2024ego}, which can serve as the visual backbone of robotic policies~\cite{r3m,pvr1,voltron,pvr2,ma2022vip,ma2023liv}. Recent works in this paradigm predict manipulation-relevant cues from web videos in the form of motion trajectories~\cite{track2act,shaw2023videodex, tao2025dexwild} and object affordances~\cite{srirama2024hrp,vrb, mendonca2023structured, shi2025zeromimic} and combine these predictive models with a limited amount of robot interaction data for training policies via conditional behavior cloning.  Our work is orthogonal to these approaches in that instead of using human videos for training, we require a single human demonstration during deployment for guiding a pre-trained generalist policy to perform a new task.

\textbf{One-Shot Imitation from Robot Demonstration.} One-shot imitation aims to enable robots to perform a new task with guidance from a single demonstration. Prior works have explored this in the context of a robot demonstration provided at test time, via training demonstration-conditioned policies~\cite{dasari2021transformers}. Recent works have developed learning-based visual servoing systems~\cite{dome} and algorithms for identifying object invariances~\cite{invariance} to replicate the robot end-effector actions from the demonstration onto novel configurations of objects. However, requiring a robot demonstration for imitation is restrictive for ubiquitous deployment as end-users might find it challenging to tele-operate the robot to collect demonstrations in natural human environments.

\textbf{One-Shot Imitation from Human Demonstration.} To imitate a human demonstration for robot manipulation, a straightforward approach is kinematic retargeting of the human hand pose to the robot end-effector pose per time-step, as demonstrated by some recent works~\cite{li2024okami,r+x, lepert2025phantomtrainingrobotsrobots}. Although simple to implement, the human-robot embodiment mismatch typically introduces errors in the retargeting and the open-loop execution is brittle to object pose variations during deployment. To remedy this, another line of works seek to learn this retargeting implicitly by training a closed-loop policy from paired human-robot datasets~\cite{jain2024vid2robot, park2025learning} enabling more flexible deployments. Since such paired datasets are difficult to collect, other works~\cite{bahl2022human, irmak, lum2025crossing, chen2024object} perform reinforcement learning with reward functions derived from video comparison. However, such methods require online interaction and resets, limiting their practicality. We propose an alternate paradigm: given the human demonstration, we refine retargeted human behavior using a pre-trained robot policy, avoiding brittle replays and enabling closed-loop control. This formulation avoids both expensive paired human-robot data collection and cumbersome test-time fine-tuning via interaction.

\section{Method}
\label{sec:method}
We target the problem of one-shot visual imitation. Given a human demonstration $\mathbf{D}$  depicting a manipulation task with description $\mathscr{T}$, we want a robot manipulator to perform the same task. Unlike prior work that relies on test-time RL training~\cite{bahl2022human} or paired human-robot datasets~\cite{jain2024vid2robot},  we wish to enable such imitation `one-shot' assuming access to a closed-loop diffusion policy $\bar{\pi}_\theta(\mathbf{a}_t | \mathbf{o}_{\leq t}, \mathcal{T})$ pre-trained on some broad robot interaction dataset $\mathcal{D}_{robot}$. Our assumption on the human demonstration $\mathbf{D}$ is also very minimal: it can either be an RGBD video or a multi-view video of a human doing a task, such that 3D hand poses per-timestep can be reliably extracted from it. 

\begin{figure*}[t]
    \centering
    \includegraphics[width=1.0\textwidth]{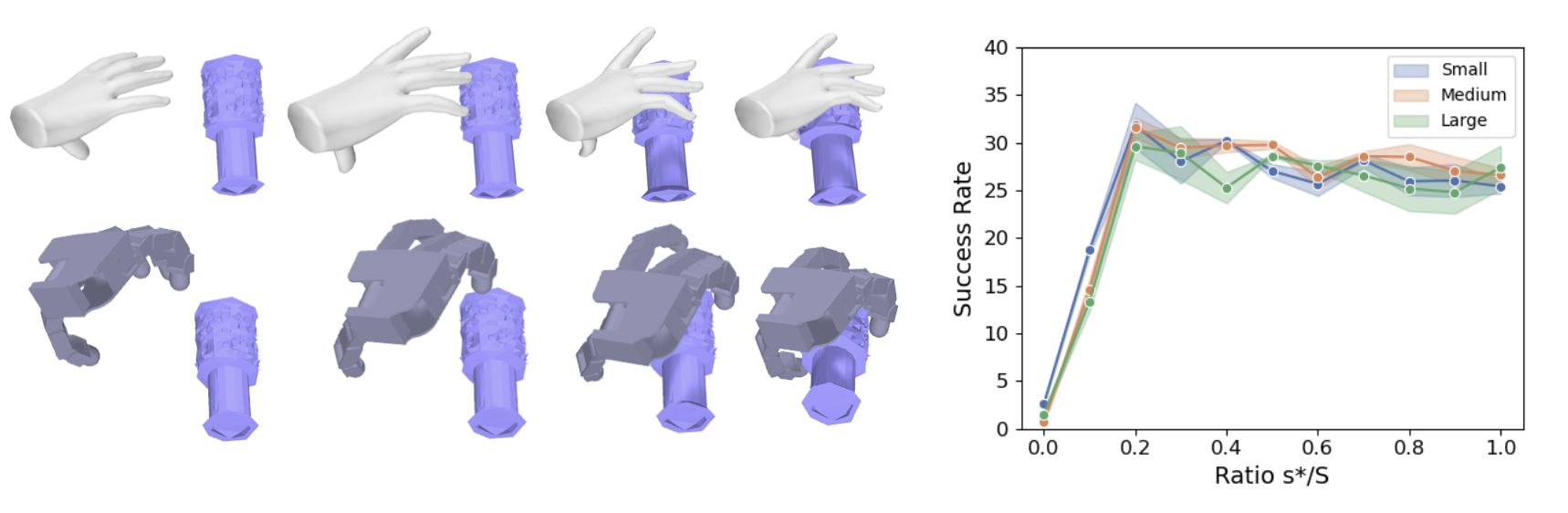}
    \caption{\textbf{Dexterous Grasping Results in Simulation.} (Left) A human demonstration and~\approach~rollout for dexterous grasping. (Right) Success rates (mean $\pm$ std over 3 seeds) as a function of diffusion step $s^*$. Here, $s^*/S=0$ corresponds to kinematic retargeting and $s^*/S=1$ corresponds to the robot policy.}
    \label{fig:simtasks}
\end{figure*}

\subsection{Overview}
Our approach is based on two insights. The first is that the trajectory of the hand pose in the human demonstration $\mathbf{D}$ provides useful information of the approximate trajectory the robot end-effector should follow, and we can perform \textit{kinematic retargeting} of the hand trajectory $\{\mathbf{h}_t\}^{T}_{t=0}$ to an open-loop robot end-effector trajectory $\{\hat{\mathbf{a}}_t\}^{T}_{t=0}$. The second insight is that the kinematically retargeted robot trajectory has the correct form of motion, but these actions may not be very precise in the distribution of plausible robot actions given the current observation. A diffusion policy models this likelihood, and we can use a pre-trained  `\textit{generalist}' diffusion policy $\bar{\pi}_\theta(\mathbf{a}_t | \mathbf{o}_{\leq t}, \mathcal{T})$ to refine the retargeted robot actions in a closed loop manner, thus inferring actions that, while still similar to the human demonstration, are more likely under the policy and better aligned with the robot embodiment (see Fig.~\ref{fig:intuition}). We describe both these steps in detail below.

\subsection{Kinematic retargeting of Human Hand Trajectories to Robot End-Effector Poses}
\label{sec:retargeting}

Given a human demonstration $\mathbf{D}$ of a manipulation task $\mathcal{T}$, our first step is to extract the 3D hand pose trajectory $\{\mathbf{h}_t\}_{t=0}^{T}$. Each hand pose $\mathbf{h}_t \in \mathbb{R}^{3 \times J}$ corresponds to the 3D locations of $J$ keypoints (e.g., wrist and fingertips), which we estimate using a pre-trained monocular hand pose estimator applied to each video frame~\cite{hamer}. These keypoints encode the motion of the human performing the task and form the basis for retargeting to a robotic end-effector.

To translate this human motion into a robot-executable trajectory, we define a simple geometric mapping function $f_{\text{retarget}}: \mathbb{R}^{3 \times J} \rightarrow \mathbb{R}^6$ that converts the human hand pose $\mathbf{h}_t$ into a robot configuration $\hat{\mathbf{a}}_t = f_{\text{retarget}}(\mathbf{h}_t)$. The mapping aligns the wrist pose of the human to the robot end-effector pose.  For a two-finger gripper, we use the distance between the thumb and the remaining fingers of the hand mesh to infer a binary robot grasp, and for a dexterous robotic hand we match the robot hand's fingertip positions to those of a human hand using inverse kinematics. We set the initial configuration of the robot as the kinematically retargeted initial configuration from the human demonstration.

This kinematic retargeting procedure yields a full trajectory $\{\hat{\mathbf{a}}_t\}_{t=0}^{T}$ in the robot's configuration space, which can be executed as an open-loop policy. However, due to differences in morphology and embodiment between humans and robots, the absence of environment feedback, and the inaccuracy of the hand estimation module, this trajectory often leads to suboptimal or unstable behavior during open-loop execution. We thus treat this as the basis for searching for a plausible robot action trajectory via denoising with a (pre-trained) closed-loop diffusion policy.

\subsection{Closed-Loop Denoising of Robot Actions with a Pre-Trained Diffusion Policy}
\label{sec:denoising}
A diffusion policy $\bar{\pi}_\theta(\mathbf{a}_t | \mathbf{o}_{\leq t}, \mathcal{T})$~\cite{chi2023diffusion, reuss2023goal} pre-trained on diverse offline robot interaction data reliably models the distribution of plausible robot actions $\mathbf{a}_t$ given previous observations $\mathbf{o}_{\leq t}$.  To predict actions $\mathbf{a}_t$, diffusion policies start with Gaussian noise $\tilde{\mathbf{a}}_t^{(S)} \sim \mathbf{N}(0, I)$ and iteratively denoise it at different diffusion steps $s$:
\begin{equation}
\tilde{\mathbf{a}}_t^{(s-1)} = \bar{\pi}_\theta\left( \tilde{\mathbf{a}}_t^{(s)}, \mathbf{o}_{\leq t} \right), \quad s = S, S-1 \dots, 0
\end{equation}

We use this to modify the kinematically retargeted trajectory  $\{\hat{\mathbf{a}}_t\}_{t=0}^{T}$ for obtaining robot actions that follow the high-level motion in the human trajectory and still lie within the distribution of plausible actions under $\bar{\pi}_\theta$. To do this, we modify this typical reverse diffusion process so that instead of starting with pure noise $\tilde{\mathbf{a}}_t^{(S)}$ at step $S$, we start from an intermediate diffusion step $s^*$  such that $0 < s^* < S$. We define  $\{\tilde{\mathbf{a}}_t^{(s^*)}\}_{t=0}^{T}$ to be a noisy version of the kinematically retargeted trajectory as follows:

\begin{equation}
\tilde{\mathbf{a}}_t^{(s^*)} = \sqrt{\alpha_{s^*}}\hat{\mathbf{a}}_t + \sqrt{1-\alpha_{s^*}}\epsilon_t, \quad \epsilon_t \sim \mathbf{N}(0, I)  
\end{equation}

Here $\alpha$ corresponds to the diffusion schedule of the pre-trained policy. This procedure, inspired by SDEdit~\cite{meng2021sdedit} which adopted a similar approach for image editing, relies on the assumption that $\{\hat{\mathbf{a}}_t\}_{t=0}^{T}$ is an approximate version of the ideal action trajectory that should be executed by the robot, and it potentially lies outside the distribution of feasible actions under the pre-trained diffusion policy $\bar{\pi}_\theta$. Intuitively, adding moderate noise to the retargeted trajectory and then denoising preserves the high-level structure of the human demonstration while allowing the policy to correct low-level infeasibilities from the embodiment gap. The diffusion policy $\bar{\pi}_\theta$ then performs iterative denoising steps, conditioned on the robot's observations $o_{\leq t}$, to refine this noisy trajectory into feasible robot actions, based on equation (2).

After $s^*$ denoising steps, the final output $\mathbf{a}_t = \tilde{\mathbf{a}}_t^{(0)}$ is deployed on the robot. 
Importantly, this process is carried out in a closed-loop manner: the policy uses real-time observations from the cameras in the scene to iteratively improve its predictions, thereby compensating for embodiment mismatch and external perturbations (e.g., object slippage or occlusion).  The key hyperparameter in this process for~\approach~is the diffusion step $s^*$ that trades off between the faithfulness to the demonstration and the likelihood under the robot policy --  in the limit $s^*=S$ we recover a rollout from the base policy $\bar{\pi}_\theta$ and in the limit $s^*=0$ we recover the kinematically retargeted trajectory $\{\hat{\mathbf{a}}_t\}_{t=0}^{T}$. While derived in the context of diffusion, note that this iterative refinement perspective also applies to flow-matching~\cite{lipman2022flow}, which we show in real-world experiments.

\begin{table}[t]
\centering
\begin{tabular}{@{}c@{\hskip 18pt}c@{\hskip 18pt}c@{\hskip 18pt}c@{}}
\toprule
\textbf{Group} & {Robot Policy} & {Kinematic} & \textbf{\approach} \\ \midrule
\textit{Small}   & 25.4 & 2.6 & \textbf{31.8} \\
\textit{Medium}  & 26.6 & 0.7 & \textbf{31.6} \\
\textit{Large}   & 27.4 & 1.5 & \textbf{29.6} \\
\midrule 
\textbf{Average} & 26.5 & 1.6 & \textbf{31.0} \\
\bottomrule
\end{tabular}
\caption{ \textbf{Simulated Dexterous Grasping Results.} The numbers show the average of success rates over 3 seeds. We use $s^*/S=0.2$, where the robot policy uses $S=1000$.~\approach~consistently outperforms both baselines, across all groups.}
\label{tab:sim}
\vspace{-0.5cm}
\end{table}

\begin{figure*}[t]
    \centering
    \includegraphics[width=\linewidth]{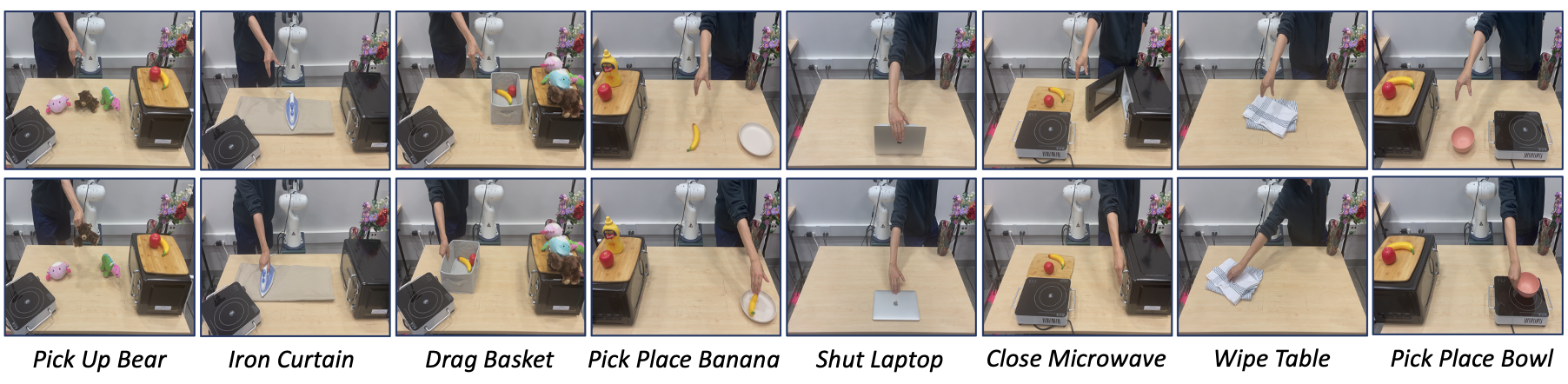}
    \caption{ \textbf{Real-World Manipulation Tasks.} Human demonstrations for the 8 evaluation tasks, shown as two frames per task. Tasks span prehensile and non-prehensile manipulation, including grasping, pushing, closing, wiping, and placing.}

    \label{fig:tasks}
\end{figure*}

\section{Experiments}
\label{sec:simexperiments}
We evaluate our approach on dexterous grasping in simulation and across diverse real-world tasks comprising prehensile and non-prehensile table-top manipulation.  Through experiments, we aim to understand the following research questions:

\begin{itemize}[]
    \item Can~\approach~outperform pure kinematic retargeting from the human demonstration? 
    \item Do the human demonstrations allow~\approach~to perform new tasks where the pre-trained diffusion policy fails?
    \item How to effectively tradeoff between faithfulness to the human demonstration and performing the task reliably, with varying noise level $s^*/S$?
    \item How robust is~\approach~to noise in human demonstration, or different kinematic retargeting?
\end{itemize}

\subsection{Dexterous Grasping in Simulation}
To verify our intuition, we first consider a simulation environment where the target task is restricted to picking up a generic object with a 16-DOF four-fingered Allegro hand. Specifically, we train a dexterous grasping policy across a small set of generic objects in simulation, and test our method on human hand grasping trajectories on a different set of objects. This serves as the pre-trained diffusion model for this experiment.

For training the robot policy, we collect a total of 985 grasping trajectories of Allegro hand over 58 training objects\cite{zhang2024graspxl} (randomly sampled 26 and 32 objects from ShapeNet~\cite{chang2015shapenet} and PartNet~\cite{mo2019partnet}, respectively), resulting in 22.2\% success rate. We use a variant of 3D Diffusion Policy~\cite{ze20243d}, which takes 3D point tracks of an object instead of point clouds. At test time, we provide human grasping trajectories on a subset of Objaverse dataset~\cite{deitke2023objaverse}, which are unseen during training. The test set contains 1220 objects, with 1 human grasping trajectory per object, provided by GraspXL~\cite{zhang2024graspxl}. Note that there is no error from hand estimation or the initial configuration of the object in this setup. We use ground-truth human hand 3D keypoints, and the object is located at the same location as the human demonstration.

\begin{figure}[h]
\centering
    \includegraphics[width=\linewidth]{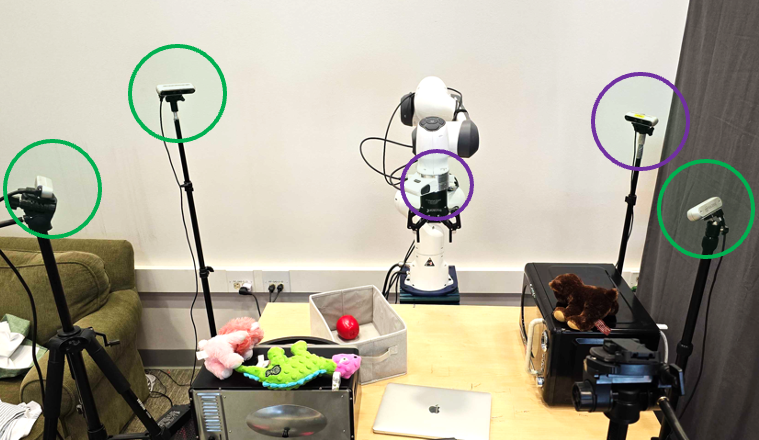}
    \caption{\textbf{Workspace with 5 Cameras.} We use the four external cameras for triangulation to obtain the global pose of the hand mesh from a human demonstration. The pre-trained policy uses the two cameras marked in purple.}
  \label{fig:setup}
\end{figure}

\textbf{Results.}
As in GraspXL~\cite{zhang2024graspxl}, we group the objects based on their size, resulting in small, medium, and large objects. We compare~\approach~with two baselines. The first is open-loop kinematic retargeting obtained from the human demonstration. The second is directly deploying the robot policy we trained. 

Table~\ref{tab:sim} shows the main results. Our method outperforms baselines, with the largest gains on small objects. We hypothesize this is because larger objects are easier to grasp, supported by base robot policy performance improving for larger objects. This experiment shows that even in a highly controlled setup where the robot environment is initialized identically to that of human demonstration, kinematic retargeting cannot solve the task, while slightly refining with a generalist grasping policy significantly improves the performance. Using kinematic retargeting to initiate the denoising process also helps the denoising process itself, outperforming the robot policy. Additionally, \approach~achieves higher inference speed, by the factor of $S/s^*$.

\textbf{Ablation studies.} We analyze the influence of $s^*$ on~\approach~in Fig.~\ref{fig:simtasks}. Overall, we observe a consistent trend of performance gain from $s^*/S=1$ to $s^*/S=0.2$, showing that~\approach~is robust to the choice of hyperparameter.

\subsection{Real-World Manipulation}
\label{sec:realexperiments}

For the real-world experiments, we use a Franka Emika Panda arm equipped with a two-finger gripper from Robotiq. We use the publicly available pre-trained flow-matching policy $\bar{\pi}_\theta(\mathbf{a}_t | \mathbf{o}_{\leq t},\mathcal{T})$ called Pi-0 released by Physical Intelligence~\cite{pi0}, trained on the DROID dataset~\cite{khazatsky2024droid}. This policy takes as input a language task specification $\mathcal{T}$, and observations from two cameras in the scene, and outputs per-timestep joint velocity. This policy was trained on a large offline dataset of tele-operated robot demonstrations, and is a \textit{generalist} policy that can perform a broad set of tasks. We do not fine-tune or adapt this policy in any way and directly use it for~\approach. We use an off-the-shelf monocular hand reconstruction model, Hamer~\cite{hamer} to obtain per-timestep hand mesh reconstructions from each of the external cameras given a human demonstration in the scene. Our experiments follow the protocol of a human first demonstrating an object manipulation in the scene, followed by robot execution after re-setting the scene.

\begin{table}[t]
\centering
\setlength{\tabcolsep}{2pt}
\renewcommand{\arraystretch}{1.1}
\begin{tabular}{cccc}
\toprule
\textbf{Task} & {Pi-0} & {Kinematic} & \textbf{\approach} \\
\midrule
Shut Down Laptop         & 20   & 10   & \textbf{60}   \\
Close Microwave     & 10   & 80   & \textbf{90}   \\
Drag Basket         & 10   & \textbf{80}   & \textbf{80}   \\
Wipe Table          & 50   & 0    & \textbf{100}  \\
Iron Curtain        & 20   & 70   & \textbf{90}   \\
Pick up Bear        & 0    & 40   & \textbf{60}   \\
Pick Place Bowl     & 0    & 70   & \textbf{100}  \\
Pick Place Banana   & 0    & 70   & \textbf{90}   \\
\midrule
\textbf{Average}       & 13.8 & 52.5 & \textbf{83.8} \\
\bottomrule
\end{tabular}
\caption{\textbf{Quantitative Results for Real-World Manipulation.} Success rates (\%) are over 10 trials per task.}
\label{tb:real}
\end{table}

\textbf{Workspace.} Fig.~\ref{fig:setup} describes our real-robot manipulation setup. The scene has 5 cameras including 4 Realsense external cameras and 1 Zed-Mini wrist camera mounted on the robot gripper. We calibrate the four external cameras and use them for triangulating the 3D hand pose so that we can obtain the position and orientation of a human hand in the world coordinates, with origin at the base of the robot. The pre-trained policy Pi-0 requires just two cameras (marked in purple) and the policy does not use any calibration information. While Pi-0 uses Zed 2 camera for external view, we use Realsense camera due to hardware availability, which introduces additional difficulty.

\begin{figure}[t]
    \centering
    \includegraphics[width=1.0\linewidth]{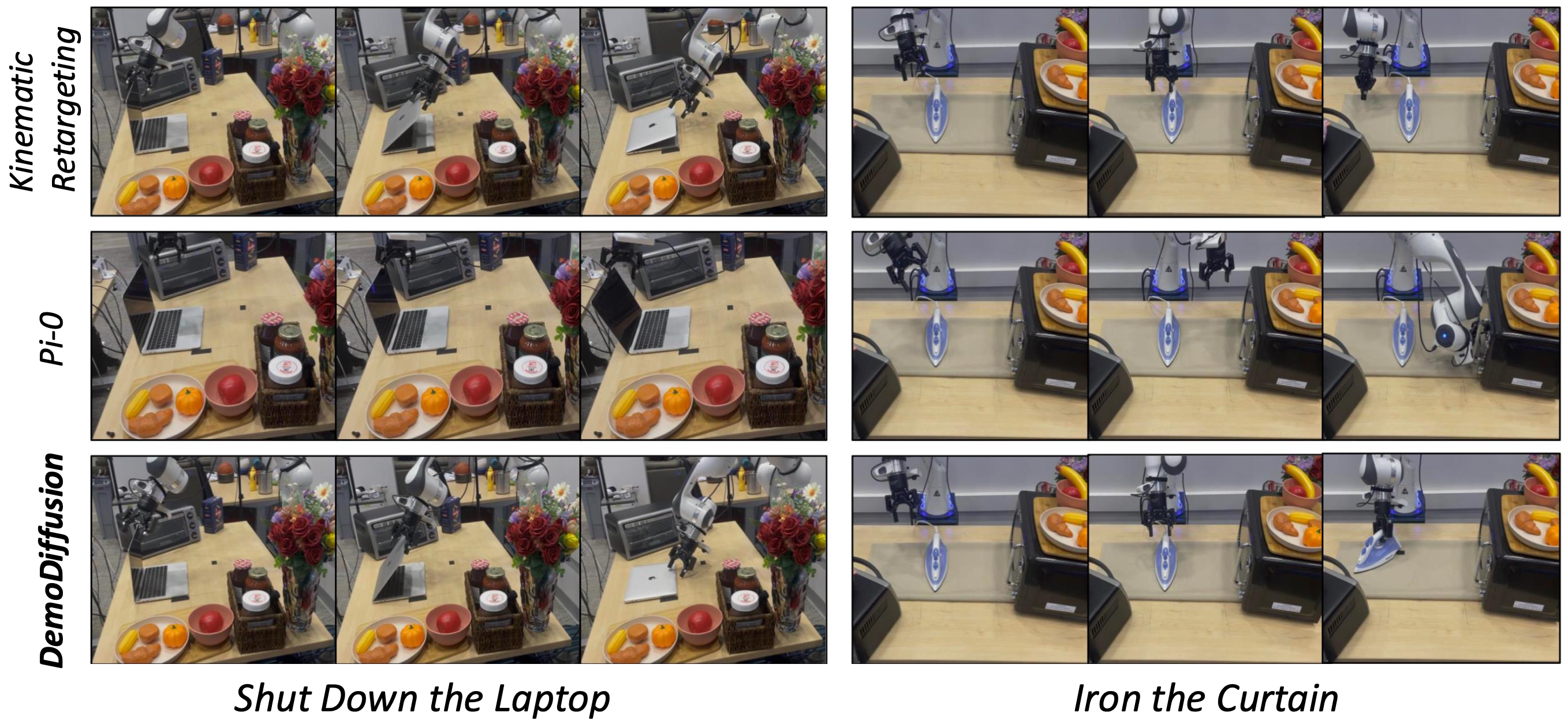}
    \caption{ \textbf{Qualitative comparisons for real-robot manipulation.} 
    Rollout progressions (start, intermediate, final frame) for two tasks. Kinematic retargeting (top) produces plausible motion but loses contact before completing the task. Pi-0 (middle) performs general reaching but fails to manipulate the correct object. \approach~(bottom) reaches the object and maintains contact through task completion. See project page for full videos.}     
    \label{fig:qual}
\end{figure}
\textbf{Tasks.} We perform evaluations for 8 different manipulation tasks, the language description and the human demonstration for which are described in Fig.~\ref{fig:tasks}. Each of these tasks requires a different manipulation strategy involving either prehensile or non-prehensile manipulation. We collect just a single human demonstration per task but for robot execution, we make organic variations in the location of objects in the scene. Additionally, we add other objects in the scene as visual distractors, to see whether \approach~can generalize under such settings. All variations are consistent across the baselines and our method.

\textbf{Baselines.} For the real-world experiments we compare~\approach~with two baselines. The first is open-loop kinematic retargeting obtained from the human demonstration. The second is directly deploying the Pi-0 model with the language instruction for the respective task. Since Pi-0 operates in the joint velocity space, for~\approach~we first convert the kinematically retargeted trajectory to the joint velocity space with IK and then add noise to the trajectory, followed by denoising with Pi-0. We deploy~\approach~with a heuristically selected noise level $s^*/S$  depending on the task -- preferring the kinematic trajectory with $s^*/S=0.2$ when Pi-0 has a zero success rate and $s^*/S=0.4$ otherwise. Note that in the limit of $0$ noise added, we recover the first baseline, kinematic retargeting, and in the limit of maximum noise $1$ we obtain the second baseline Pi-0.

\textbf{Qualitative Visualizations.} In Fig.~\ref{fig:qual}, we show rollouts for two tasks across all the methods.  We observe that Pi-0 can only perform general reaching behaviors: it either fails to follow the manipulation trajectory needed for the task or confuses the target object with a distractor. On the other hand, we observe that (open-loop) kinematic retargeting from the human demonstration has a plausible motion of the robot end-effector, but it completely misses grasping the iron, and loses contact with the laptop before fully closing it. Compared to these,~\approach~reaches the respective object and also follows the approximate motion of the retargeted trajectory while remaining in contact with the object to solve the tasks.

\begin{table}[t]
\centering

\setlength{\tabcolsep}{4pt}
\renewcommand{\arraystretch}{1.1}
\begin{tabular}{ccc}
\toprule
         & {Shut Laptop} & {Pick Place Banana} \\
\midrule
\textbf{\approach} & {60} & {90}\\
+5cm noise & {40} & {70}\\
with Thumb+Index & {30} & {80}\\
\midrule
Thumb+Index & 0 & 30 \\
\midrule
\end{tabular}
\caption{ \textbf{Robustness of~\approach.} We ablate~\approach~under noisy 3D keypoints (2nd row) and with different kinematic retargeting (3rd row). For noisy 3D keypoints, we randomly shift the tracked hand 3D keypoints by 5cm. For different kinematic retargeting, we use the distance between the thumb tip and the index fingertip instead of all fingers.}
\label{tb:real_abl_obj}
\vspace{-0.1cm}
\end{table}

\textbf{Quantitative Comparisons.} In Table~\ref{tb:real}, we compare~\approach~and the baselines across all the real-world manipulation tasks. \approach~matches or outperforms both baselines across all 8 tasks, with the largest gains on tasks requiring precise contact -- e.g., 60\% vs. 10--20\% on shutting the laptop and 100\% vs. 0--50\% on wiping the table. Notably, even for tasks where both baselines fail,~\approach~succeeds by operating at an intermediate noise level $s^*$ that combines the trajectory structure from the human demonstration with the policy's closed-loop refinement.

\textbf{Noisy Human Demonstrations.} To evaluate the robustness of ~\approach~to imperfect inputs, we injected noise into the tracked human trajectories and evaluated~\approach. Specifically, we place the object at the same location as the human demonstration, and randomly shift the tracked hand 3D keypoints by 5cm. The results in Table~\ref{tb:real_abl_obj} indicate that ~\approach~maintains high-quality performance despite the presence of tracking noise, thereby reducing the reliance on highly accurate 3D annotations in human demonstration.

\textbf{Different Kinematic Retargeting.} Lastly, we further assessed the sensitivity of our method to the choice of retargeting by evaluating an alternative approach (`Thumb+Index'). Specifically, we use the distance between thumb and index finger~\cite{lepert2025phantomtrainingrobotsrobots}, instead of all fingers, to infer the grasp. Table~\ref{tb:real_abl_obj} shows the results. While this alternative retargeting itself shows degraded performance compared to the previous retargeting,~\approach~consistently improves task performance despite suboptimal retargeting.





\section{Discussion}
\label{sec:discussion}

We presented \approach,  a simple approach to leverage \textit{generalist} diffusion models for imitating a human demonstration. By retargeting hand motion from the human demonstration into an approximate robot end-effector trajectory and modifying it through de-noising with a pre-trained diffusion policy, \approach~enables the robot to produce actions that are both aligned with the demonstrated intent and contextually plausible. Our method consistently outperforms both the base diffusion policy and the raw retargeted trajectory across diverse tasks in simulation and the real world, without requiring additional training, paired demonstrations, or reward functions. We envision that this can serve as a starting point for future efforts at human imitation across generic scenarios and perhaps also yield an  improved exploration strategy for methods pursuing policy adaptation with online RL.

\section*{Limitations}
\label{sec:limit}
Our approach has several limitations. First, it assumes that the robot should act similarly to the human to successfully complete the task, which may not hold in scenarios requiring different strategies due to embodiment or environmental constraints. Second, while effective for one-shot imitation (not requiring training / test-time fine-tuning), the method does not produce a reusable task policy that can generalize across arbitrary variations of the task. Third, the quality of the retargeted trajectory is crucial: accurate 3D human motion capture is challenging and errors in retargeting can impact downstream performance, though our method shows a degree of robustness to such noise. Additionally, the method implicitly assumes that the timing and speed of human and robot actions are aligned; extending the approach to allow for temporal alignment at test time is a promising direction of future work.

\section*{Acknowledgements}
\label{sec:ack}
We appreciate the helpful discussions with Yanbo Xu, Qitao Zhao, and Lucas Wu. We would also like to thank Kenny Shaw, Tony Tao, Jiahui Yang, Andrew Wang, Jason Liu, Hengkai Pan, Mohan Kumar Srirama, Peiyuan Yang for helping set up the hardware. This work was supported by gift awards from CISCO, Google, and NSF Award IIS-2442282.





\bibliographystyle{unsrt}
\bibliography{bib/reference}
\clearpage
\appendix
We provide additional experimental details, implementation details, and results for the real-world and simulation experiments.

\section*{Real-World Manipulation}
\label{sec:sup_real}
\subsection{Hardware Setup}
We replicate DROID~\cite{khazatsky2024droid} setup for the hardware setup. Due to hardware availability, we use RealSense D455 instead of ZED 2 for the rear view camera.

\subsection{Task Setup Description}
\begin{figure*}[t]
    \centering
    \includegraphics[width=0.9\textwidth]{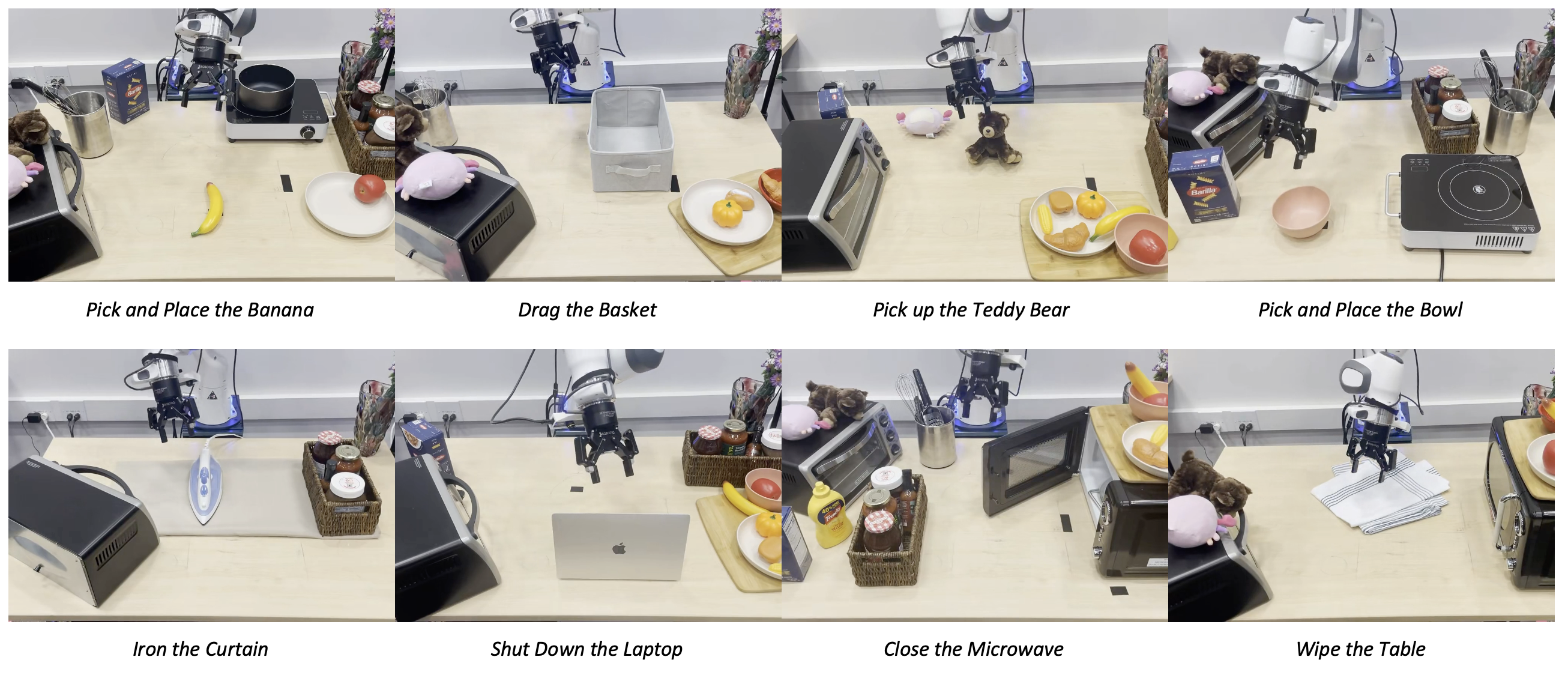}
    \caption{\footnotesize \textbf{Real-World Manipulation Tasks at Test Time.} Each image corresponds to an example of the scene for each task at test time. To test our method in a natural setup, we add other objects in the scene as visual distractors. For each rollout, we use the same set of distractors across methods.}
    \label{fig:realtest}
\end{figure*}

The overview of scenes for each task at test time is shown in Fig.~\ref{fig:realtest}. During data collection, we mark the target object location with a tape. During evaluation, we randomize the target object location within the tape, which is consistent across all methods. Overall, kinematic retargeting may fail due to small misalignment of object placement, inaccurate 3D hand keypoints, or camera extrinsics, etc. Pi-0 may fail if it cannot identify the object to interact with and manipulate other objects in the scene. It may also reach the target object but fail to manipulate in the desired way. Note that previous approaches~\cite{li2024okami, r+x, wang2024dexcap, chen2024arcap} use kinematic retargeting to transfer human demonstration to the robot.

\textbf{Pick and Place the Banana.}
The task is to pick up the toy banana and place on the plate on the right side of the table. The episode is considered successful if the banana is successfully placed on the plate.

\textbf{Drag the Basket.}
The task is to drag the basket to the right side of the table. The human demonstration grasps the edge of the basket and pulls it to the right. The episode is considered successful if the robot grasps the basket/ puts the gripper inside the basket and move it to any side of the table.

\textbf{Pick Up the Teddy Bear.}
The task is to pick up the teddy bear located at the center of the table. The human demonstration grasps the neck of the bear and picks it up. The episode is considered successful if the robot picks up the bear by the end of the episode.

\textbf{Pick and Place the Bowl.}
The task is to pick up the bowl and place on the stove on the right side of the table. The episode is considered successful if the bowl is successfully placed on the stove.

\textbf{Iron Curtain.}
The task is to grasp the ironing machine located at the center of the table and slide it to the right side of the curtain. The episode is considered successful if the robot grasps/pushes the ironing machine to any part of the curtain.

\textbf{Shut Down the Laptop.}
The task is to close the laptop located in front of the robot. The episode is considered successful if the angle of the laptop is smaller than 10 degrees at the end of the episode.

\textbf{Close the Microwave.}
The task is to close the microwave located at the corner of the table. The episode is considered successful if the handle part touches the body part of the microwave at the end of the episode.

\textbf{Wipe the Table.}
The task is to pinch/poke the given tissue and wipe the table. The episode is considered successful if the robot pinches/pokes the tissue and performs one stroke to any side of the table without missing it.

\subsection{Text Prompts}
We use the following text conditions for Pi-0~\cite{pi0}, which are also used for \approach.

\begin{table}[h]
\centering
\small
\renewcommand{\arraystretch}{1.25}
\caption{Text prompts used for real-world manipulation tasks.}
\label{tab:text_prompts}
\begin{tabular}{p{0.42\columnwidth} p{0.50\columnwidth}}
\toprule
\textbf{Task} & \textbf{Prompt} \\
\midrule
Pick and Place the Banana &
\textit{pick and place the banana on the plate} \\

Drag the Basket &
\textit{drag the basket to the right} \\

Pick Up the Teddy Bear &
\textit{pick up the bear doll} \\

Pick and Place the Bowl &
\textit{pick and place the bowl on the stove} \\

Iron the Curtain &
\textit{move the ironing machine to the right} \\

Wipe the Table &
\textit{wipe the table} \\

Close the Microwave &
\textit{close the microwave} \\

Shut Down the Laptop &
\textit{close the laptop} \\
\bottomrule
\end{tabular}
\end{table}

\subsection{Inference Details}
At each timestep $t$, we index the human demonstration at same timestep $t$, and compute the kinematically retargeted actions. As Pi-0 predicts a chunk of actions with length $H=10$, we also use the action chunk computed from human demonstration from timestep $t$ to $t+H-1$. The episode length is set to be the same as the length(number of timesteps) of human demonstrations across all methods.

\subsection{Kinematic Retargeting Implementation Details}
A human demonstration $\mathbf{D}$ of a manipulation task $\mathcal{T}$ consists of the 3D hand pose trajectory $\{\mathbf{h}_t\}_{t=0}^{T}$. Each hand pose $\mathbf{h}_t \in \mathbb{R}^{3 \times J}$, which corresponds to the 3D locations of $J$ keypoints (e.g., wrist and fingertips). To kinematically retarget these 3D keypoints to a robotic end-effector, we use the human's wrist position $A$, thumb tip position $B$, and the average of other fingertips $C$. Specifically, $\vec{AB}$, along with the cross product of $\vec{AB}$ and $\vec{AC}$ define the orientation of the end effector, and the human's wrist position is used as the position of the end effector. As Pi-0 uses joint velocity as action space, the end effector pose is further converted into joint velocity using inverse kinematics. For the gripper, we use the norm of $\vec{BC}$ as its width. To stabilize grasping, we set the gripper action to be fully closed if the norm of $\vec{BC}$ is smaller than $80\%$ of the gripper's maximum width, similar to~\cite{lepert2025phantomtrainingrobotsrobots}.

\subsection{Hyperparameters}
To show our method's applicability across diverse setups, we do not change the speed of the given human demonstration when performing kinematic retargeting. At inference, we use the following hyperparameters for Pi-0 and \approach, with only difference on the number of denoising steps, as~\approach~starts denoising from intermediate steps. Specifically, we use the following hyperparameters. Depending on the noise level $s^*/S$, \approach~uses $20\times s^*/S$ denoising steps.

\begin{table}[t]
\renewcommand{\arraystretch}{1.3} 
\centering
\caption{Hyperparameters in Real-world.}
\begin{tabular}{@{} l c @{}} 
\toprule
\textbf{Hyperparameter} & \textbf{Value} \\
\midrule
Open Loop Horizon & 8 \\
Predict Action Horizon & 10 \\
Denoising Steps at Inference (Pi-0) & 20 \\
\bottomrule
\end{tabular}
\label{tab:additional_hyperparameters}
\vspace{-0.5cm}
\end{table}

\subsection{Additional Results}

\textbf{Results with Different Hyperparameters.} We present the results of \approach~using two different hyperparameter settings, $s^*/S = 0.2$ and $s^*/S = 0.4$. The results are summarized in Table~\ref{tb:real_add}. Overall, we observe a consistent trend based on the performance of the base diffusion policy (\textit{Pi-0}):

\begin{itemize}
    \item When the base policy \textit{Pi-0} achieves a non-zero success rate, $s^*/S = 0.4$ generally yields better performance. This is observed in tasks such as \textit{shut laptop}, \textit{close microwave}, \textit{drag basket}, \textit{wipe table}, and \textit{iron curtain}.
    \item Conversely, when \textit{Pi-0} has a zero success rate, $s^*/S = 0.2$ tends to perform better. This is the case for tasks including \textit{pick up bear}, \textit{pick place banana}, and \textit{pick place bowl}. We observe that while $s^*/S = 0.4$ shows adaptive end effector motion, it tends to avoid activating the gripper at the correct timestep. 
\end{itemize}

\noindent
These findings suggest that a more effective base policy enables \approach{} to benefit from a higher noise level, as it can leverage the base policy’s strengths more effectively. In contrast, when the base policy is less reliable, a lower noise level tends to show better performance. We believe this can serve as a guideline for choosing the noise level $s^*/S$ for~\approach.

\begin{figure}[tbp]
    \centering
    \includegraphics[width=1.0\linewidth]{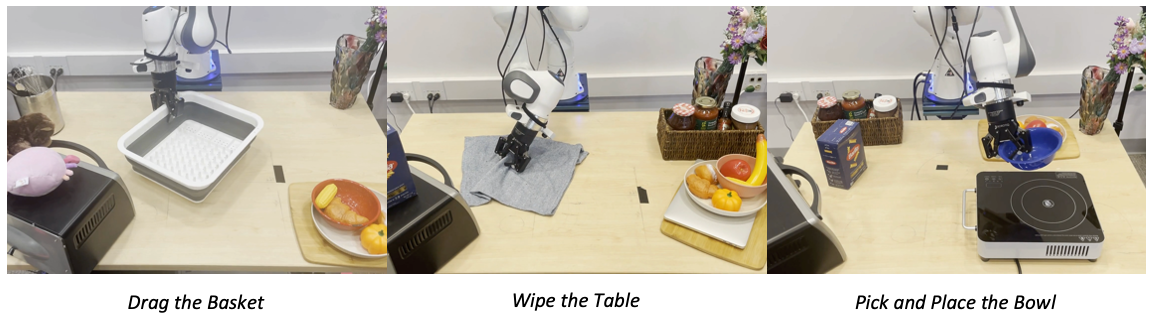}
    \caption{\footnotesize \textbf{Zero-Shot Generalization to New Objects.} We evaluate~\approach's generalization capabilities to new objects on 3 tasks - drag the basket, wipe the table, pick and place the bowl. While the objects have different colors and shapes from those of objects in the human demonstration,~\approach~adjusts its behavior to these new objects. Please refer to the website for detailed visualizations.}
    \label{fig:generalization}
\end{figure}

\textbf{Zero-Shot Generalization to New Objects.} 
Lastly, we investigate whether~\approach~can achieve zero-shot generalization to new objects within the same category. We use the same human demonstration and text prompt we previously used while evaluating on a new object within the same category. Results are shown in Fig.~\ref{fig:generalization}. Although we only provide human demonstration on a different object, we observe that~\approach~can adapt its behavior to a new object.

\section*{Dexterous Grasping in Simulation}
\label{sec:sup_sim}
\subsection{Task Setup Description}
In our simulation experiment, the task is to pick up the object. The rollout is considered success when the object's z position is higher than 0.10 \textit{m} at the end of the episode. Similar to the real-world experiments, at each timestep of inference, we compute a kinematically retargeted action chunk computed from human demonstration at the same timestep.

\textbf{Training Data.} As we don't have a \textit{generalist} pre-trained diffusion policy in simulation, we first train a diffusion policy while constraining the target task to picking up generic objects. Specifically, we collected total 985 grasping trajectories of Allegro hand 
over 58 objects in the RaiSim simulator for training. The trajectories were generated by rolling out the expert Allegro RL policy with random grasping directions. The 58 objects are randomly sampled from ShapeNet~\cite{chang2015shapenet} (26 objects) and PartNet~\cite{mo2019partnet} (32 objects). The training objects and expert Allegro RL policy are provided by GraspXL\cite{zhang2024graspxl}.

\begin{table}[tbp]
\centering
\setlength{\tabcolsep}{1.5pt} 
\renewcommand{\arraystretch}{1.1} 
\caption{\textbf{Additional Quantitative Results for Real-World Manipulation.}}
\resizebox{\columnwidth}{!}{ 
\begin{tabular}{lcccc}
\toprule
\textbf{Task} & \textbf{Pi-0} & \textbf{\begin{tabular}[c]{@{}c@{}}Kinematic\\ Retargeting\end{tabular}} & \textbf{\begin{tabular}[c]{@{}c@{}}\approach\\ ($0.2$)\end{tabular}} & \textbf{\begin{tabular}[c]{@{}c@{}}\approach\\ ($0.4$)\end{tabular}} \\
\midrule
Shut Laptop         & 20 & 10 & 40 & \textbf{60} \\
Close Microwave     & 10 & 80 & 80 & \textbf{90} \\
Drag Basket         & 10 & \textbf{80} & 70 & \textbf{80} \\
Wipe Table          & 50 & 0  & 60 & \textbf{100} \\
Iron Curtain        & 20 & 70 & 70 & \textbf{90} \\
Pick Up Bear        & 0  & 40 & \textbf{60} & 40 \\
Pick Place Banana   & 0  & 70 & \textbf{90} & 20 \\
Pick Place Bowl     & 0  & 70 & \textbf{100} & 0 \\
\midrule
\textbf{Average}    & 13.8 & 52.5 & \textbf{71.3} & 60.0 \\
\bottomrule
\end{tabular}
}
\label{tb:real_add}
\end{table}

\textbf{Test Data.} At test time, we provide human grasping trajectories for a different set of objects. As it is challenging to provide human demonstration in simulation, we instead use MANO~\cite{romero2022embodied} hand RL policy to collect human demonstrations. The test set contains total 1220 objects, which is a subset of Objvaerse dataset~\cite{deitke2023objaverse}. The objects are divided in three groups based on its size: small scale $s\in[3,5]cm$ medium scale  $m\in[5,7]cm$, and large scale $l\in[7,9]cm$. As the object meshes don't contain material information, we assume fixed density, resulting in diverse mass. We additionally use the same friction coefficient for all objects. Again, test objects and expert MANO~\cite{romero2022embodied} RL policy are provided by GraspXL\cite{zhang2024graspxl}. Additionally, we do not change the speed of the given human demonstration when performing kinematic retargeting.

\subsection{Kinematic Retargeting Implementation Details}
Following~\cite{wang2024dexcap}, we use inverse kinematics to match the fingertip positions between the human hand and the robot hand, ignoring the pinky finger. The pose of the robot hand base is initialized to be the same as that of a human hand.

\subsection{Robot Policy Implementation Details}
We train a hierarchical policy, where the high-level planner predicts the robot's future goal state, and the low-level controller predicts the action given the desired future robot state. Both policies were trained using a variant of 3D Diffusion Policy~\cite{ze20243d}, where we modify the point cloud encoder to take point tracks, for a richer representation of the object. Specifically, while the original point encoder takes object points at each timestep, the modified encoder takes a sequence of object points (object tracks) as input.

High-level planner and low-level controller use a subset of the following as input and output. All rotations are represented in Euler angles. 

\textbf{Robot state.} contains its wrist position in the world frame, rotation in the world frame, and absolute hand joint angles. 

\textbf{Object state.} contains 3D points of its surface. For a history of robot state, we assume 3D point tracks to be given(i.e. each 3D point across timesteps corresponds to the same point of the object).

\textbf{Robot goal state.} contains its wrist position in the local frame(wrist frame at previous timestep), rotation in the local frame, and relative hand joint angles to the previous timestep. 

\textbf{Action.} contains the desired robot's wrist position in the local frame(previous wrist frame), desired rotation in the local frame, and desired relative hand joint angles to the previous timestep. Note that the action and the robot's goal state are not the same. For instance, the desired hand joint angles can penetrate the object surface, while the robot's state cannot. Such a difference creates contact force, enabling a stable grasp.

\textbf{High-level Planner.}
The high-level planner takes in a history of robot states and object states, and predicts a sequence of robot goal states. At test-time, we use the kinematic retargeting result as an initial estimate for the robot's goal state. We then apply \approach~to refine the robot goal state, and feed it to the low-level controller.

\textbf{Low-level Controller.}
The low-level controller takes in a history of robot states and a sequence of robot goal states, and predicts a sequence of robot actions. The goal state encoder has the same architecture to robot state encoder. Once it outputs the action, the action is used as a position target for a PD controller provided in the RaiSim simulator.

\subsection{Hyperparameters}
For policy training, we use the default hyperparameters provided in the 3D Diffusion Policy~\cite{ze20243d} with minimal modifications. For the simulation environment, we use the default hyperparameters provided by GraspXL~\cite{zhang2024graspxl}. Specifically, we use the following same parameters for both the high-level planner and low-level controller.

\begin{table}[h!]
\renewcommand{\arraystretch}{1.3} 
\centering
\caption{Hyperparameters in Simulation.}
\begin{tabular}{@{} l c @{}} 
\toprule
\textbf{Hyperparameter} & \textbf{Value} \\
\midrule
${H}$ & 4 \\
${N_{obs}}$ & 2 \\
$N_{act}$ & 1 \\
$N_{goal}$ & 2 \\
$N_{latency}$ & 0 \\
Number of Object Points  & 1024 \\
Train Timesteps & 1000 \\
Inference Timesteps & 10 \\
Episode Length & 155 \\
Wrist Position Pgain & 100.0 \\
Wrist Position Dgain & 0.1 \\
Wrist Rotation Pgain & 100.0 \\
Wrist Rotation Dgain & 0.2 \\
Finger Rotation Pgain & 100.0 \\
Finger Rotation Dgain & 0.2 \\
\bottomrule
\end{tabular}
\label{tab:hyperparameters}
\end{table}

\end{document}